\documentclass{article} % For LaTeX2e
\usepackage{iclr2025_conference,times}
\usepackage{amsmath,amsfonts,bm}

\def\eqref#1{equation~\ref{#1}}
\def\1{\bm{1}}

\DeclareMathAlphabet{\mathsfit}{\encodingdefault}{\sfdefault}{m}{sl}
\SetMathAlphabet{\mathsfit}{bold}{\encodingdefault}{\sfdefault}{bx}{n}

\usepackage{booktabs}
\usepackage{hyperref}
\usepackage{url}
\hypersetup{
  colorlinks=true,
  linkcolor=black,
  citecolor=black,
  urlcolor=black
}

\title{Gated Prototype Transfer for Zero-Support Medical Vision Language Model}

\author{
\begin{tabular}{@{}p{0.47\textwidth}@{}p{0.47\textwidth}@{}}
\normalfont
\textbf{Sumaiya Fazal}\thanks{Corresponding author.} \newline
FAST-NUCES \newline
\texttt{p200401@nu.edu.pk}
&
\normalfont
\textbf{Rizwan Qureshi} \newline
Massachusetts General Hospital, \newline
Harvard University \newline
\texttt{rizwan.qureshi@mgh.harvard.edu}
\\[1.5em]
\normalfont
\textbf{Laleh Behjat} \newline
University of Calgary \newline
\texttt{laleh@ucalgary.ca}
&
\normalfont
\textbf{Samira Ebrahimi Kahou} \newline
University of Calgary/Mila \newline
\texttt{samira.ebrahimikahou@ucalgary.ca}
\end{tabular}
}

\newcommand{\method}{GProtoT}
\newcommand{\gated}{Gated-ProtoT}
\newcommand{\sstext}{SS-Text+}
\newcommand{\mcr}{\mathrm{MCR}}

\newcommand{\clipval}[1]{\operatorname{clip}_{[0,1]}\!\left(#1\right)}
\iclrfinalcopy
\begin{document}

\maketitle
% \vspace{-1em}
% \begin{center}
%     \textbf{Accepted to SMASH 2026}
% \end{center}
% \vspace{1em}

\begin{abstract}
Realistic few-shot adaptation of medical vision-language models (VLMs) can leave low-prevalence classes without any labeled support. The training-free \sstext{} solver remains operational in this setting, but its zero-support prototypes receive no visual adaptation and fall back to a purely textual class prototype. We introduce Gated Knowledge-Guided Prototype Transfer (\method{}), a pre-repulsion module that transfers supported-class visual residuals to missing classes using VLM similarity and an ontology-based clinical distance. A validation-free evidence gate scales transfer based on semantic relevance and residual agreement, thereby limiting unsupported extrapolation. In 50 paired realistic one-shot trials on MESSIDOR with FLAIR features, severe and proliferative diabetic retinopathy were absent from every support set. Relative to \sstext{}, gated ProtoT improved missing-class recall by 5.23 points, average class accuracy by 1.37 points, and the missing/supported recall harmonic mean by 1.83 points. It reduced negative-transfer frequency from 46-48\% for ungated transfer to 2\%, while at the cost of reductions of 1.20 percentage points in supported-class recall and 0.45 percentage points in top-1 accuracy. These results identify that evidence-based gating provides a safety mechanism for the knowledge source. \method{} is based on \sstext{} and requires no gradients, validation data, or additional model inference, offering a low-compute path toward accountable adaptation under incomplete clinical supervision. Code is available: \href{https://github.com/sumaiyaafridi/Gated-KGBT}{https://github.com/sumaiyaafridi/GProtoT}
\end{abstract}

\section{Introduction}

Medical VLMs provide transferable representations and text-derived class prototypes, making black-box few-shot adaptation possible without updating a large encoder \citep{shakeri2024fewshot}. Yet the conventional balanced $K$-shot protocol is poorly aligned with disease prevalence. Common findings may appear repeatedly while rare or severe categories receive no labeled examples, and aggregate accuracy can then remain high even when clinically important categories are unrecognized.

\citet{silvarodriguez2025fewshot} introduced \sstext{}, a closed-form text-based linear probe to formalize realistic, imbalanced, validation-free adaptation. For a class with no support, \sstext{} correctly retains its text prototype rather than fitting an undefined visual evidence. This behavior is robust, but limited-supported classes become task-adapted while a missing class remains purely zero-shot. An alternate method is to borrow a related supported prototype. Our experiments show that borrowing a related supported prototype strategy is unsafe and, due to ungated VLM and ontology-guided transfer, reduced balanced performance and missing-class recall is nearly half of support trials.

We therefore ask whether cross-class transfer can improve a zero-support prototype without turning external knowledge into a source of negative transfer? We propose evidence-based \method{}, a training-free module applied after \sstext{} forms class prototypes and before its standard repulsion step. The main formulation transfers visual residuals, preserving the missing class's text semantics, and the evidence-based gate shrinks the update when semantic relevance or source agreement is weak. Our contributions are:
\begin{itemize}
    \item a non-recursive residual-transfer rule that employs only originally supported classes
    \item a VLM, ontology, and hybrid neighbor construction with a validation-free evidence-based gate 
    \item a paired safety evaluation that reports missing-class recall, supported-class recall, their harmonic mean, and negative-transfer frequency, not only overall accuracy
\end{itemize}
\section{Methodology}
\subsection{Knowledge Guided Prototype Transfer}

Building on \sstext{}, let $t_c\in\mathbb{R}^D$ be the text prototype of class $c$, $K_c$ its support count, and $\tilde{\mu}_c$ its pre-repulsion \sstext{} prototype. Class-adaptive regularization gives $\tilde{\mu}_c=t_c$ when $K_c=0$ \citep{silvarodriguez2025fewshot}. Thus, supported classes receive visual adaptation, whereas zero-support classes remain text-only. We define the supported and missing sets as $\mathcal{S}=\{i:K_i>0\}$ and $\mathcal{M}=\{c:K_c=0\}$, respectively.

For Knowledge-guided neighbors, for each missing target $c^*\in\mathcal{M}$, ProtoT selects sources only from $\mathcal{S}$; updated missing classes are never reused as sources. Candidate sources are ranked using text-space VLM similarity,
$s^{\mathrm{vlm}}_{c^*,i}=\cos(t_{c^*},t_i)$, ontology proximity,
$s^{\mathrm{ont}}_{c^*,i}=-d(c^*,i)$, or the average of their z-scored values for hybrid ranking. After selecting the top-$k$ neighbors $\mathcal{N}(c^*)$, their scores are converted into normalized contribution weights:
\begin{equation}
w_i=
\frac{\exp(s_i/T)}
{\sum_{j\in\mathcal{N}(c^*)}\exp(s_j/T)},
\qquad
\sum_{i\in\mathcal{N}(c^*)}w_i=1.
\label{eq:ProtoT-weights}
\end{equation}
\eqref{eq:ProtoT-weights} assigns greater influence to more relevant sources while keeping the total contribution controlled. For a non-positive user temperature, ProtoT uses the validation-free value
$T=\max(\operatorname{std}\{s_i\},10^{-3})$.

For residual transfer and evidence gate, rather than copying complete source prototypes, ProtoT transfers only their task-specific visual corrections. With $\hat{x}=x/\lVert x\rVert_2$, the residual of supported class $i$ is
$r_i=\hat{\tilde{\mu}}_i-\hat{t}_i$. The weights in \eqref{eq:ProtoT-weights} then produce the proposed target correction $\bar r_{c^*}$, while $a_{c^*}$ measures whether the selected residuals agree:
\begin{equation}
\bar r_{c^*}
=
\sum_{i\in\mathcal{N}(c^*)}w_i r_i,
\qquad
a_{c^*}
=
\frac{\lVert\bar r_{c^*}\rVert_2}
{\sum_{i\in\mathcal{N}(c^*)}w_i\lVert r_i\rVert_2+\epsilon}.
\label{eq:ProtoT-residual}
\end{equation}
Aligned residuals give $a_{c^*}\approx1$, whereas conflicting residuals cancel and reduce agreement. Because agreement alone does not establish target relevance, ProtoT also computes $q_{c^*}$: VLM relevance is the clipped nearest-neighbor margin above the candidate-row median, ontology relevance is $1/\max(d_{\min},1)$, and hybrid relevance is their geometric mean. The gate and effective strength are
$g_{c^*}=\clipval{a_{c^*}q_{c^*}}$ and
$\beta_{c^*}=\beta g_{c^*}$.

The missing prototype is updated by adding the gated residual to its original text direction:
\begin{equation}
p_{c^*}
=
\lVert t_{c^*}\rVert_2
\frac{\hat{t}_{c^*}+\beta_{c^*}\bar r_{c^*}}
{\lVert\hat{t}_{c^*}+\beta_{c^*}\bar r_{c^*}\rVert_2}.
\label{eq:ProtoT-update}
\end{equation}

This preserves the target’s text semantics while incorporating only evidence-supported visual information. If evidence is weak, $g_{c^*}=0$ and ~\eqref{eq:ProtoT-update} recovers the original text prototype. ProtoT leaves supported prototypes unchanged, and the original \sstext{} repulsion and inference normalization proceed as before. It requires no gradients, validation data, extra labels, or additional VLM inference, preserving the efficiency of training-free black-box adaptation \citep{schwartz2020greenai}.

\section{Experimental Setup}

We build upon the official \sstext{} implementation and utilize its cached FLAIR representations for grading diabetic retinopathy in MESSIDOR \citep{decenciere2014messidor, silvarodriguez2025flair}. The classification includes five ordinal categories: no retinopathy, mild, moderate, severe, and proliferative diabetic retinopathy. Following the realistic K = 1 protocol, the total dataset for support consists of five images sampled according to the training prevalence. Across seeds 0–49, each support set contained counts of $[3,1,1,0,0]$, while the identities of support images varied; severe and proliferative disease categories were consistently absent.  Prevalence-weighted sampling at $k=1$ yields the same class count pattern in every seed, so seed variance measures only sensitivity to support selection. Because the identity of the absent classes is not varied, missing-class results should be interpreted as specific to this particular zero-support configuration.

All techniques utilize the same support sets and initial prototypes. ProtoT employs residual transfer with parameters set to $k=3$, $\beta=1$, automatic temperature control, and the original all-class repulsion. Both ProtoT-VLM and ProtoT-Ont operate without gating. \method{} incorporates hybrid ranking along with the proposed gating mechanism. Each experiment is conducted alongside \sstext{} using the identical support set. The ontology is a manually specified ordinal class graph based on the established diabetic-retinopathy severity ordering \citep{wilkinson2003proposed}.

Because typical instances predominate in the test set, top-1 accuracy is insufficient. In light of the focus on class-aware medical validation \citep{maierhein2024metrics}, we therefore present average class accuracy (ACA), mean recall over missing classes (MCR), mean recall over supported classes (SCR), and harmonic mean ($H$).

\section{Results}

\begin{table}[t]
\caption{Fifty-seed realistic one-shot MESSIDOR performance. Parentheses report paired changes from \sstext{} in percentage points. NTR and severe NTR denote the percentages of seeds with $\Delta\mcr<0$ and $\Delta\mcr<-5$ points, respectively}
\label{tab:results}
\centering
\small
\setlength{\tabcolsep}{3.2pt}
\begin{tabular}{lrrrrrrr}
\toprule
Method & Top-1 & ACA & MCR & SCR & $H$ & NTR & Sev. \\
\midrule
\sstext{} & 57.45 & 54.64 & 54.90 & 54.46 & 52.87 & -- & -- \\
ProtoT-VLM & 54.83 & 51.91 & 52.73 & 51.37 & 50.25 & 46 & 38 \\
 & $(-2.62)$ & $(-2.72)$ & $(-2.16)$ & $(-3.10)$ & $(-2.63)$ & & \\
ProtoT-Ont & 54.92 & 51.90 & 52.47 & 51.52 & 50.20 & 48 & 34 \\
 & $(-2.53)$ & $(-2.74)$ & $(-2.43)$ & $(-2.94)$ & $(-2.67)$ & & \\
\gated{} & 57.00 & \textbf{56.01} & \textbf{60.12} & 53.27 & \textbf{54.71} & \textbf{2} & \textbf{2} \\
 & $(-0.45)$ & $(+1.37)$ & $(+5.23)$ & $(-1.20)$ & $(+1.83)$ & & \\
\bottomrule
\end{tabular}
\end{table}

Table~\ref{tab:results} shows that knowledge transfer alone does not guarantee safe transfer. Relative to \sstext{}, both ungated variants yield lower mean performance across all reported metrics. ProtoT-VLM reduces ACA, MCR, SCR, $H$, and top-1 accuracy by 2.72, 2.16, 3.10, 2.63, and 2.62 percentage points, respectively and the corresponding reductions for ProtoT-Ont are 2.74, 2.43, 2.94, 2.67, and 2.53 points. Negative transfer is frequent where ProtoT-VLM and ProtoT-Ont reduce MCR relative to the paired baseline in 46\% and 48\% of seeds, respectively, with reductions exceeding five points in 38\% and 34\% of seeds. For ProtoT-VLM, the mean MCR change is negative but statistically inconclusive, mean change $=-2.16$ points, 95\% CI $[-4.73,\,0.32]$, Holm-adjusted $p=.103$, because the interval includes zero. In contrast, ProtoT-Ont produces a statistically significant, although borderline, reduction in MCR, mean change $=-2.43$ points, 95\% CI $[-4.68,\,-0.21]$, Holm-adjusted $p=.049$.\footnote{Holm correction is applied across the three planned MCR comparisons to control the family-wise error rate arising from multiple paired tests.}

\gated{} reverses the missing-class pattern. It raises MCR to 60.12\%, corresponding to a mean paired improvement of 5.23 percentage points, 95\% CI $[3.84,\,6.69]$. It also increases ACA to 56.01\%, $+1.37$ points, 95\% CI $[0.78,\,1.99]$ and $H$ to 54.71\%, $+1.83$ points, 95\% CI $[1.06,\,2.64]$. MCR improves in 36 of the 50 seeds and remains unchanged in 13, leaving only one seed with negative transfer, NTR $=2\%$. These gains are accompanied by modest reductions of 1.20 points in SCR, 95\% CI $[-1.52,\,-0.88]$ and 0.45 points in top-1 accuracy, 95\% CI $[-0.71,\,-0.19]$. Thus, the combined hybrid-ranking and evidence-gating variant substantially improves recognition of zero-support classes.

Moderate DR is the highest-weighted source for both missing classes across all 50 seeds. The mean gate is $0.582\pm0.010$ for severe DR and $0.214\pm0.003$ for proliferative DR. The weaker proliferative update reflects its larger nearest-supported ontology distance of two and lower VLM relevance of $0.102$, compared with distance one and relevance $0.381$ for severe DR. Nevertheless, gate magnitude controls the strength of the prototype update rather than directly determining final recall, which also depends on the baseline prototype, class separability, and resulting decision boundaries. Under \gated{}, per-class recall is 46.53\% for severe DR and 73.71\% for proliferative DR, compared with 37.47\% and 68.00\% under ProtoT-VLM and 36.93\% and 68.00\% under ProtoT-Ont. 

\begin{table}[t]
\centering
\small
\setlength{\tabcolsep}{4pt}
\caption{Skin histology results with CONCH \citep{lu2024visual} over 50 paired realistic
nominal $K=1$ seeds. Performance values are mean percentages;
$\Delta$ reports paired changes from \sstext{} in percentage points.
NTR and Sev. denote the percentages of seeds with
$\Delta\mathrm{MCR}<0$ and $\Delta\mathrm{MCR}<-5$ points, respectively.}
\label{tab:skin_results}
\begin{tabular}{lrrrrrrr}
\toprule
Method & Top-1 & ACA & MCR & SCR & $H$ & NTR & Sev. \\
\midrule
\sstext{} & 77.577 & 57.166 & 12.276 & \textbf{84.100}
          & 21.344 & -- & -- \\
\gated{}  & \textbf{79.444} & \textbf{63.193}
          & \textbf{28.709} & 83.884
          & \textbf{42.729} & 0 & 0 \\
$\Delta$ (points) & $+1.867$ & $+6.027$ & $+16.433$
                 & $-0.216$ & $+21.385$ & -- & -- \\
\bottomrule
\end{tabular}
\end{table}

Table~\ref{tab:skin_results} extends the evaluation to Skin histology
with CONCH \citep{lu2024visual} under 50 paired realistic nominal $K=1$ seeds. We retain
the MESSIDOR transfer settings: hybrid ranking, $k=3$, $\beta=1$,
automatic temperature, residual transfer, and all-class repulsion.
Every seed contains zero-support classes. Relative to \sstext{},
\gated{} raises MCR from 12.276\% to 28.709\%, a gain of 16.433
percentage points, and improves ACA and $H$ by 6.027 and 21.385
points, respectively. Top-1 accuracy increases by 1.867 points,
while SCR decreases by 0.216 points. No seed has lower MCR than
its paired baseline (NTR $=0/50=0\%$), and severe negative transfer
is also absent. Both Skin and MESSIDOR show improved missing-class
recall, although their top-1 trade-offs differ: Skin gains 1.867
points, whereas MESSIDOR loses 0.45 points. These observations
extend the combined variant to a non-ordinal histology label space
and a second encoder.

\section{Conclusion}

We introduced knowledge-guided prototype transfer under a realistic one-shot protocol in which prevalence-weighted support allocation leaves the classes with no samples entirely unsupported. Knowledge proximity alone did not ensure safe transfer: both VLM and ontology ungated variants lowered every mean performance metric and produced negative transfer in 46 to 48\% of the support-sampling seeds. Evidence gating substantially reduced this failure. \gated{} improved mean missing-class recall by 5.23 percentage points and ACA by 1.37 points, while reducing negative transfer to one of 50 seeds. These improvements were accompanied by modest observed reductions of 1.20 points in supported-class recall and 0.45 points in top-1 accuracy.

The results suggest that calibrating transfer strength can reduce the risks of knowledge-guided transfer. However, because GProtoT combines hybrid ranking with evidence gating, the present comparison does not fully isolate the gate's
contribution. Skin histology with CONCH \citep{lu2024visual} further shows a 16.43-point MCR gain and a 1.87-point top-1 gain, with no negative MCR changes in 50 paired seeds. These findings extend the evaluation beyond retinal grading, while broader class-rotated validation and isolating the gate’s contribution remain necessary.

\bibliography{iclr2025_conference}
\bibliographystyle{iclr2025_conference}

\end{document}